\documentclass[11pt]{article}
\usepackage{geometry}
\usepackage{coling2020}
\usepackage{times}
\usepackage{url}
\usepackage{latexsym}
\usepackage{microtype}
\usepackage{latexsym}
\usepackage{graphicx}
\usepackage{caption, subcaption}
\usepackage{url}
\usepackage{amsmath}
\usepackage[export]{adjustbox}

\colingfinalcopy 

\title{SemEval-2020 Task 10:\\Emphasis Selection for Written Text in Visual Media}

\author{Amirreza Shirani$^{\dagger}$, 
Franck Dernoncourt$^{\ddagger}$,
Nedim Lipka$^{\ddagger}$,\\
\textbf{ Paul Asente$^{\ddagger}$,
Jose Echevarria$^{\ddagger}$,
and Thamar Solorio$^{\dagger}$} \\
 	$^{\dagger}$University of Houston \qquad
 	$^{\ddagger}$Adobe Research \qquad \\
 	$^{\dagger}$\small{\texttt{\{ashirani,tsolorio\}@uh.edu}} \\
 	$^{\ddagger}$\small{\texttt{\{franck.dernoncourt,lipka,asente,echevarr\}@adobe.com}} \\
 	}

\date{}

\begin{document}
\maketitle
\begin{abstract}
In this paper, we present the main findings and compare the results of SemEval-2020 Task 10, Emphasis Selection for Written Text in Visual Media. 
The goal of this shared task is to design automatic methods for emphasis selection, i.e. choosing candidates for emphasis in textual content to enable automated design assistance in authoring.
The main focus is on short text instances for social media, with a variety of examples, from social media posts to inspirational quotes. 
Participants were asked to model emphasis using plain text with no additional context from the user or other design considerations.
SemEval-2020 Emphasis Selection shared task attracted 197 participants in the early phase and a total of 31 teams made submissions to this task. 
The highest-ranked submission achieved 0.823 $\mbox{Match}_m$ score. 
The analysis of systems submitted to the task indicates that BERT and RoBERTa were the most common choice of pre-trained models used, and part of speech tag (POS) was the most useful feature. Full results can be found on the
task’s website\footnote{\url{https://competitions.codalab.org/competitions/20815}}.
\end{abstract}

\section{Introduction}
\label{intro}
%
%
\blfootnote{
    %
    %
    %
    %
    %
    %
    \hspace{-0.65cm}  
    This work is licensed under a Creative Commons 
    Attribution 4.0 International License.
    License details:
    \url{http://creativecommons.org/licenses/by/4.0/}.
}

%

In visual communication, emphasis is an intentional focus on one or more components to create a main focal point or center of interest with the composition. It has been found that it only takes the human eye 50 milliseconds to form an opinion on a visual composition \cite{lindgaard2006attention}. Therefore, it is important for a visual element to deliver a clear message by calling attention to specific information. Whether on flyers, posters, ads, social media posts or motivational messages, emphasis is usually designed to grab a viewer’s attention by being distinct from the rest of the design elements. 

\begin{figure}[t]
\centering
\begin{subfigure}{0.30\textwidth}
\includegraphics[width=\linewidth, left]{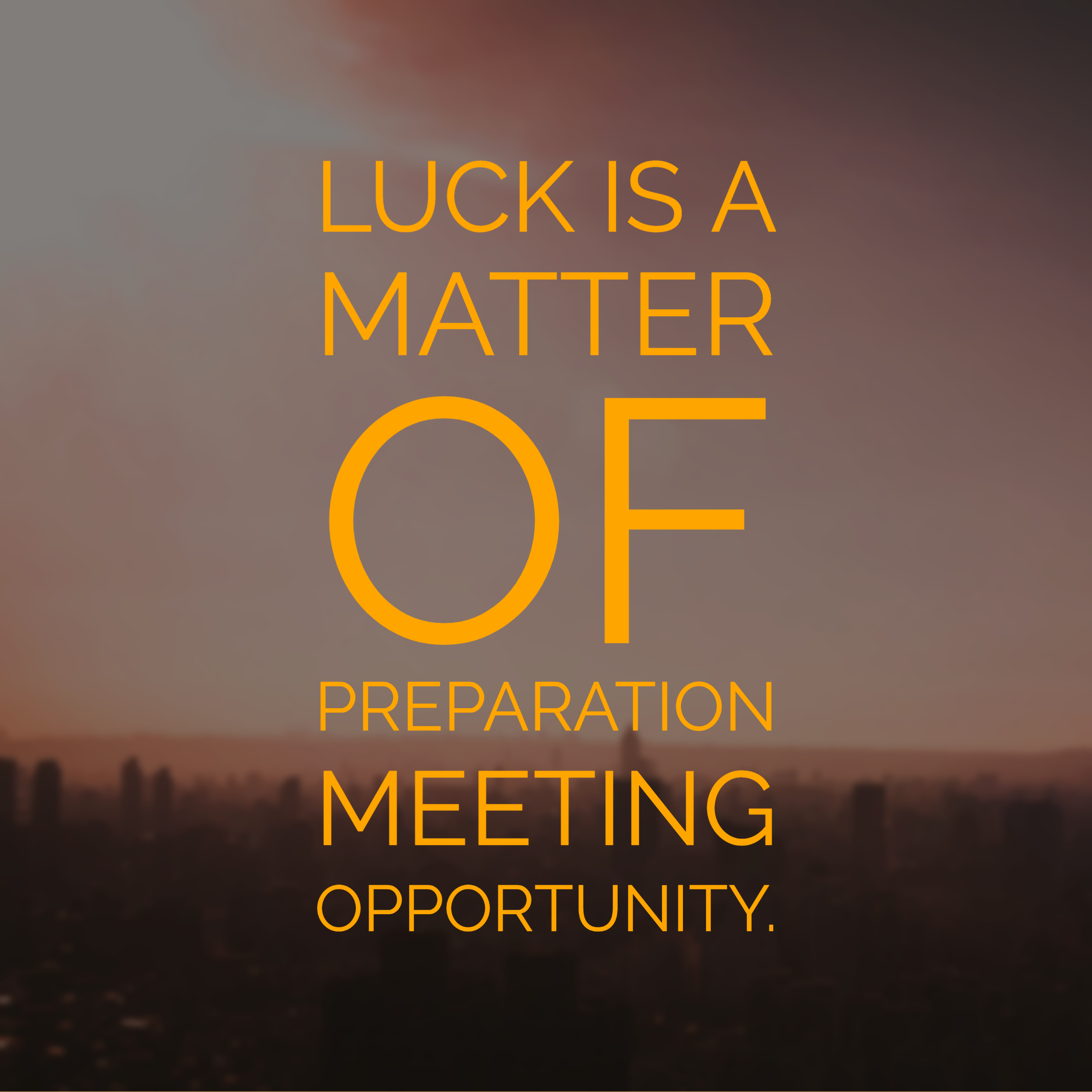} 
\caption{}
\label{fig:expa}
\end{subfigure}
\begin{subfigure}{0.30\textwidth}
\includegraphics[width=\linewidth, right]{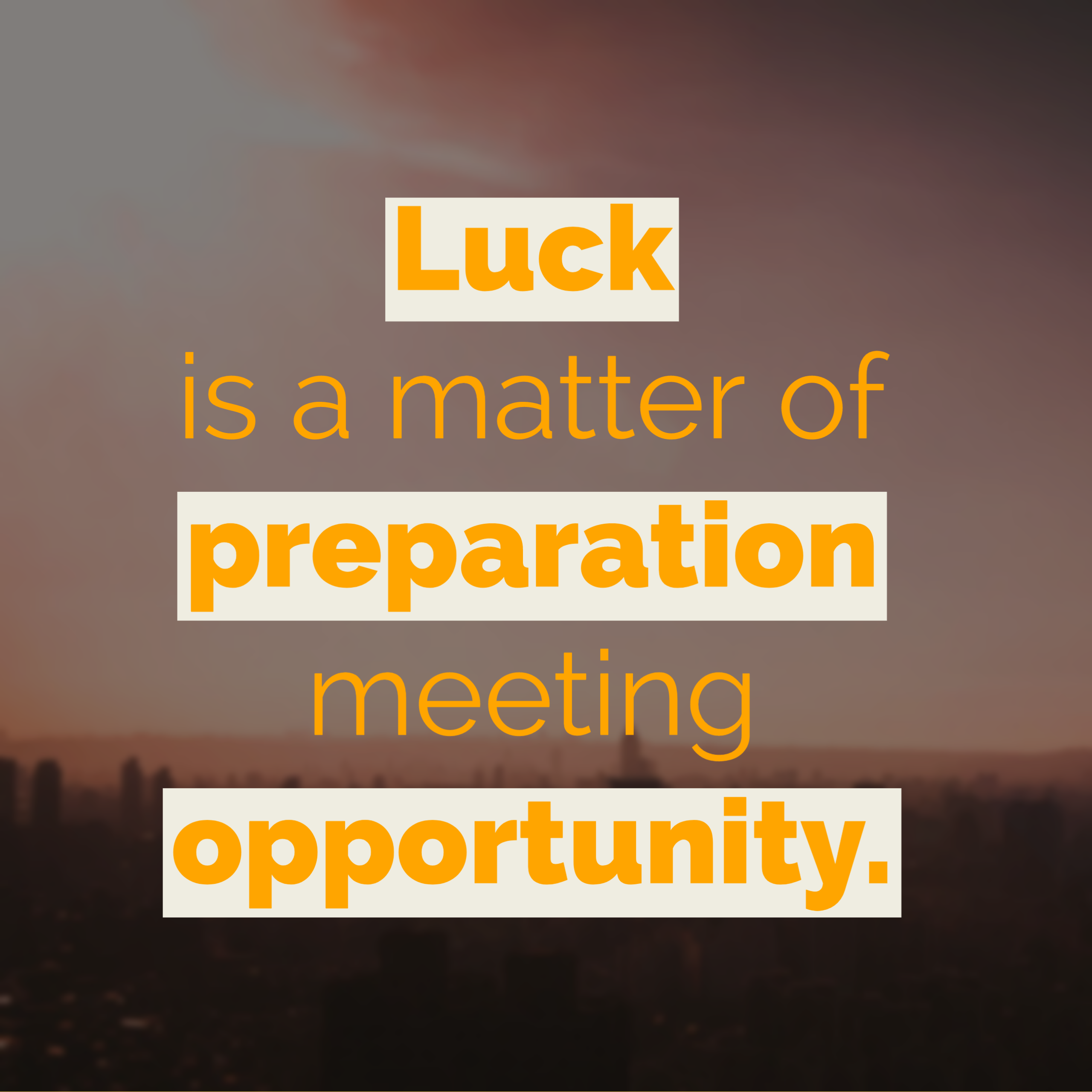}
\caption{}
\label{fig:expb}
\end{subfigure}
\caption{\small{Two different text layouts emphasizing different parts of the sentence.}}
\label{fig:image2}
\end{figure}

Thanks to various online platforms, a massive amount of digital text is being generated by users every day. 
These media are filled with content competing for attention and are usually highly designed to engage viewers' attention to convey their message in the most efficient way.
For textual content, word emphasis is used as a powerful tool to better convey the desired meaning of the written text to the audience. Utilizing emphasis techniques can potentially add another dimension to the text through visualization.  
Emphasis on textual content can be done with colors, backgrounds, or fonts, or with styles like italic and boldface to clarify or even change the meaning of a sentence by drawing attention to some specific information.
Figure~\ref{fig:expa} shows an example that is aesthetically appealing but fails to effectively communicate its intent. Understanding the text would allow the system to propose a different layout that emphasizes words that contribute more to the communication of the intent, as shown in Figure~\ref{fig:expb}.

In the last few years, we have observed many significant improvements in various platforms for generating, formatting, and editing digital text. For example, some graphic design applications such as Adobe Spark\footnote{\url{https://spark.adobe.com}} perform automatic text emphasis using templates that include images and text with different design effects. However, the used layout algorithms are often inflexible in that they rigidly emphasize words based on the visual attributes (e.g., word length) of those words, rather than the semantics of the text. 
As a result, the outcome may fail to accurately communicate the meaning of the written text, resulting in unintended emphasis and the wrong message to the audience.
However, an emphasis selection model can potentially make better suggestions by having a better understanding of the input text.

\paragraph{Task Characteristics } 
This emphasis selection task poses new challenges associated with the nature of the task: (1) No additional context from the user or the rest of the design such as background image is provided. 
Therefore the proposed task requires a computational understanding of the written text. 
(2) The dataset contains very short texts, usually fewer than ten words. Generally, working with short text instances is challenging since the decision needs to be made by only considering a few words. 
(3) Word emphasis patterns are author- and domain-specific, therefore, without knowing the author’s intent and only considering the input text, multiple emphasis selections are valid. However, a good model should be able to capture the inter-subjectivity or common sense within given annotations and finally label words according to higher agreements.

\paragraph{Expected Impact of the Task }
The ultimate goal is to enable design assistance for authors by suggesting words that are good candidates to emphasize. 
The typical applications of this task include, but are not limited to, creating flyers, posters, drawings, advertisements and other
visual material one may find online and across social media platforms such as Pinterest, Instagram and Snapchat. Moreover, emphasis selection models have applications in many design programs such as Adobe Spark, Apache OpenOffice Impress, GIMP, or Microsoft PowerPoint.

\paragraph{SemEval Emphasis Selection Task } In this shared task, we invited research on a novel Natural Language Processing (NLP) task that represents unique algorithmic and modeling challenges due to its nature. 
We observed a diverse and interesting set of solutions to tackle the existing challenges from a large number of participants, both from academia and industry. As part of this shared task, we released a dataset annotated with word emphases, which served as a benchmark to compare various techniques. Furthermore, we expect the task to be interesting for researchers studying relevant tasks such as machine-human interaction, reading comprehension, graphic design and user experience.
In the following sections, we describe the setup, participation, results, and more importantly, the insights gained from the task.

\subsection{Task Definition} 
Given a sequence of tokens $C=\{x_1,...,x_n\}$, a real number $y_i\in[0,1]$ needs to be assigned for each token in the sequence, indicating the degree to which the token needs to be emphasized. 
In other words, we define the emphasis score $y_i$ as the probability or weight of the $i^{\text{th}}$ token in the sequence. 
Finally, during the evaluation, the final set of emphases are generated by selecting tokens with the highest values (described in Section~\ref{sec:eval}).


\section{Related Work}
We firstly introduced and formulated the task of emphasis selection in \cite{shirani2019learning} in which an end-to-end label distribution learning (LDL) model in a sequence tagging architecture is proposed to model emphasis. We evaluated the model against different baselines on the Spark dataset (introduced in Section~\ref{sec:data}). 

Keyword or key-phrase detection may be the closest topic to emphasis selection. 
Keywords can capture the main topics described in a given document \cite{turney2002learning}. Modeling keywords or key-phrases has been widely addressed in different domains such as news articles \cite{wan2007towards}, scientific publications \cite{nguyen2007keyphrase} and Twitter data \cite{zhang2016keyphrase,bellaachia2012ne}. 
Keyword detection mainly focuses on finding important nouns or noun phrases \cite{augenstein2017semeval}.
In contrast, emphasis could be applied to a subset of words with different roles in a sentence. Generally, word emphasis may use to express emotions, show contrast, capture a reader’s interest or clarify a message. Moreover, emphasis selection in social media posts deals with very short texts and the prediction needs to be made based on a single instance.

In the context of expressive prosody generation, emphasis has been addressed based on acoustic and prosodic features that exist in spoken data. For example, \cite{nakajima2014emphasized} predicted emphasized accent phrases from advertisement text information and \cite{mass2018word} modeled word emphasis on audience-addressed speeches.

%

\section{Data Collection} \label{sec:data}
The data used for this shared task is the integration of two datasets from different sources, which are created from scratch based on texts collected from the Adobe Spark and Wisdom Quotes website. The dataset used for this task can be found in the task's data repository\footnote{\url{https://github.com/RiTUAL-UH/SemEval2020_Task10_Emphasis_Selection}}. The following are the descriptions of the two datasets.
The \textbf{Spark dataset} is collection of 1,195 instances from Adobe Spark\footnote{https://spark.adobe.com}. It contains a variety of subjects featured in flyers, posters, advertisements or motivational memes on social media. 
The \textbf{Quotes dataset} is a collection of quotes from well-known authors collected from Wisdom Quotes\footnote{http://wisdomquotes.com} with 2,681 instances. 

\begin{figure}[h]
\centering
\includegraphics[width=0.60\linewidth]{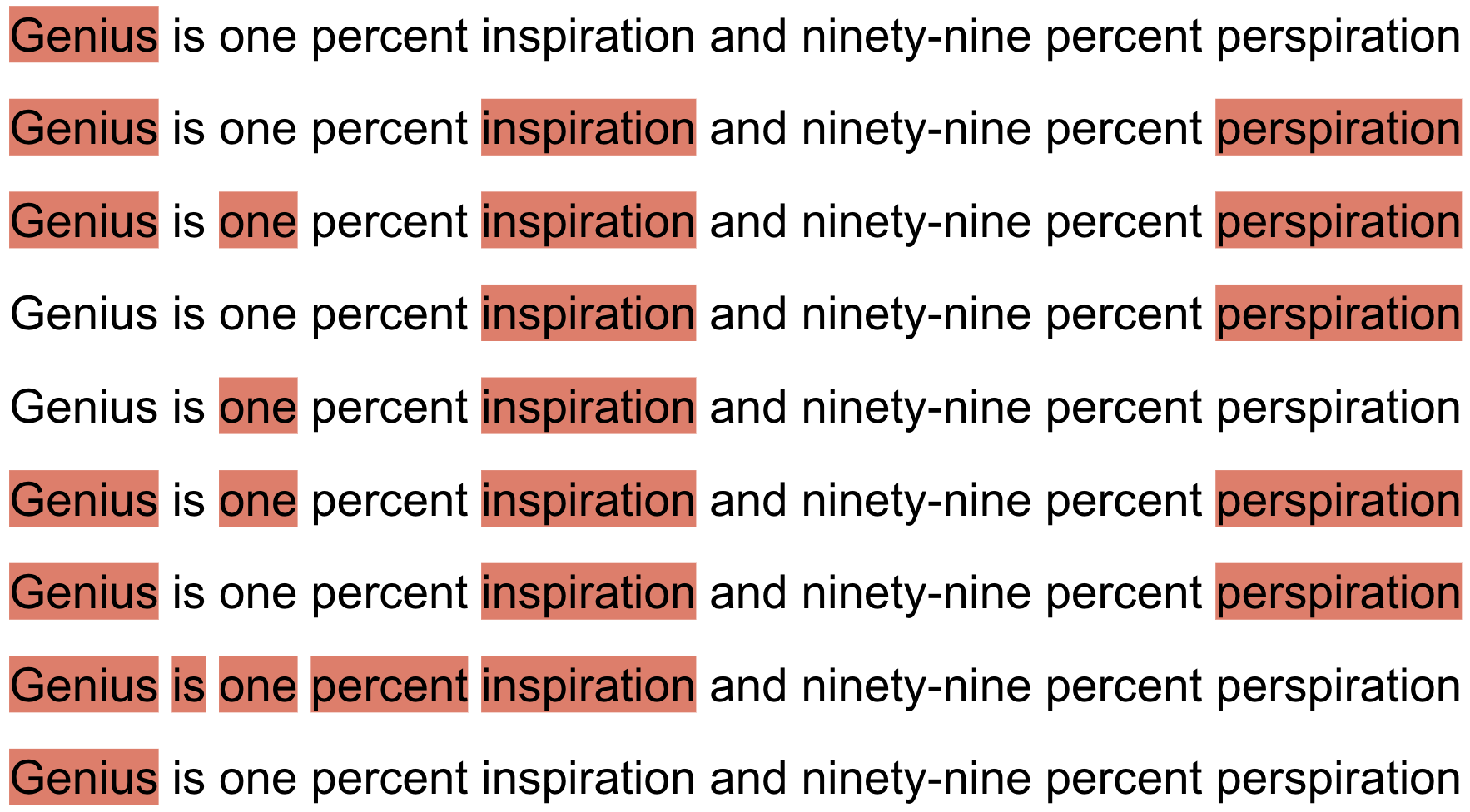} 
\caption{A short text example from the Quotes dataset along with its nine annotations.}
\label{fig:anntExample}
\end{figure}

Table~\ref{tab:stats} provides details about the length of instances in the datasets. The Emphasis dataset with 3,876 instances, consists of  44,976 words and 4,886 unique words.
We used Amazon Mechanical Turk and asked nine annotators to label each piece of text.
More precisely, we asked annotators to select word(s) in the given text that should be emphasized. Having nine annotators gives us this ability to capture different viewpoints, each focusing on different parts of the sentence. Figure \ref{fig:anntExample} shows an example of text annotated with nine annotations. 
In this example, there is more consensus in emphasizing words like ``inspiration" and ``Genius". On the other hand, words like ``is" and ``percent" are not good candidates based on general agreement. 
To ensure high-quality annotation, we included carefully-designed quality questions in 10 percent of the hits. Moreover, we only allowed master annotators to participate. 
 
The data is split up randomly between training, development and test sets.
A training data set of 2,741 instances, development set of 392 instances, and test set of 743 instances were released to the participants.   
Fleiss’ Kappa score \cite{shrout1979intraclass} of 24.60 was observed on the data set. Such a Kappa score indicates the
existence of multiple perspectives about emphasis in the dataset.

\begin{table}[h]
\centering
\caption{Dataset Statistics}
\label{tab:stats}
\resizebox{0.47\textwidth}{!}{%
\begin{tabular}{|l|ccc|}
\hline
Dataset/Text length &  \multicolumn{1}{l|}{Max} & \multicolumn{1}{l|}{Avg.} & \multicolumn{1}{l|}{SD} \\ \hline
Spark dataset       &  29                       & 6.24                       & 4.64                     \\ \cline{1-1}
Quotes dataset      &  38                       & 13.99                      & 5.47  
\\ \cline{1-1}
Emphasis dataset (combined)      &  38                       & 11.60                      & 6.33
\\ \hline
\end{tabular}%
}
\end{table}
%

Table~\ref{tab:label} shows an example of a short text annotated with the BIO annotations. As it is shown, words such as ``Best" are selected more often for emphasis than other words in the sequence. 
First, we compute the label distribution for each instance, which corresponds to the count per label normalized by the total number of annotations (shown in ``Norm. Freq. column"). Then we compute \emph{Emphasis Probabilities} for all the words in the sequence. The final evaluation is against ground truth emphasis probabilities (explained in Section~\ref{sec:eval}).

\begin{table*}[hb]
\centering
\caption{An example from the
Spark dataset along with its nine annotations.
In this table,``B/I"s and ``O"s represent emphasis and non-emphasis words respectively. ``B"s indicate the beginning and ``I"s indicate the inside of emphasis. ``Freq." and ``Norm. Freq." columns show the normal and normalized values for label frequencies respectively.}
\label{tab:label}
\resizebox{\textwidth}{!}{%
\begin{tabular}{|c|ccccccccc|c|c|c|}
\hline
\textbf{Words} & \textbf{A1} & \textbf{A2} & \textbf{A3} & \textbf{A4} & \textbf{A5} & \textbf{A6} & \textbf{A7} & \textbf{A8} & \textbf{A9} & \textbf{Freq. {[}B,I,O{]}}& \textbf{Norm. Freq. {[}B,I,O{]}} & \textbf{Emphasis Probs {[}B+I{]}}\\ \hline
Tag & B & O & B & O & B & O & O & O & B & {[}4,0,5{]} & {[}0.44,0,0.55{]} & {[}0.44{]}\\
Your & O & O & O & O & O & O & O & O & O & {[}0,0,9{]} & {[}0,0,1{]} & {[}0{]}\\
Best & O & B & B & B & B & B & O & B & B & {[}7,0,2{]} & {[}0.77,0,0.22{]} & {[}0.77{]}\\
Friends & O & I & O & O & O & I & B & O & I & {[}1,3,5{]} & {[}0.11,0.33,0.55{]} & {[}0.44{]}\\
\hline
\end{tabular}%
}
\end{table*}

\section{Data Analysis}
Many systems reported performance gain by using Part of Speech Tags (POS) tags in their models. In this section, we analyze the effectiveness of this feature by closely examining the top 20 POS tags in our dataset. We used the Stanford Part-Of-Speech Tagger \cite{toutanova2003feature} to obtain POS tags for all tokens in our dataset.  
We divide the emphasis probabilities to four intervals (0-0.25, 0.25-0.50, 0.50-0.75 and 0.75-1.00) and compute how the POS tags are distributed in these four intervals.

\begin{figure}[h]
\centering
\includegraphics[width=0.70\linewidth]{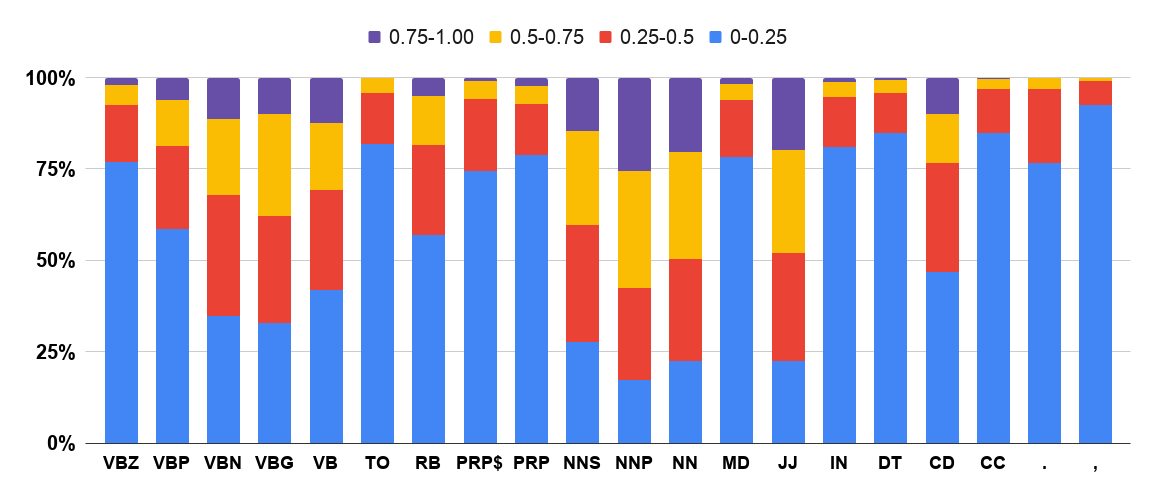} 
\caption{Frequencies of the top 20 POS tags in 0-0.25, 0.25-0.5, 0.5-0.75, 0.75-1.00 intervals of emphasis probabilities.
The vertical values correspond to the percentage of tag counts over the total number of words in training set.}
\label{fig:pos}
\end{figure}

Figure \ref{fig:pos} shows the occurrence of the top 20 POS tags in four emphasis probability intervals for all token labels in our training set. POS tags like ``,", ``.", ``DT" and ``PRP" are more favored to have low emphasis probabilities (0-0.25). Interestingly, words with the highest probabilities (0.75-1.00) are usually from ``NNP", ``NN" and ``JJ" word types. 
As we expected, there are some general trends for emphasized words with respect to the type of words in sentences, which make POS tags a useful feature for modeling emphasis.

\section{Evaluation Metric}
\label{sec:eval}
The evaluation was performed on the test set. Participants were asked to provide a real value (greater and equal to zero) for each token in the test set that indicates the probability of the token being emphasized. All models were evaluated with $\mbox{Match}_m$ metric and ranked based on the averaged values of scores for m=${1,2,3,4}$.
\paragraph{$\mbox{Match}_m$} For each instance $x$ in the test set $D_{test}$, we selected a set $S^{(x)}_m$ of $m \in \{1 \dots 4\}$ words with the top $m$ probabilities according to the ground truth. Analogously, we selected a prediction set $\hat{S}^{(x)}_m$ for each $m$, based on the predicted probabilities.
We defined the metric $\mbox{Match}_m$ as follows:
\[\mbox{Match}_m := \frac{\sum_{x \in D_{test}} |S^{(x)}_m \cap \hat{S}^{(x)}_m|/m}{|D_{test}|}\]

Finally, we computed the average value of $\mbox{Match}_m$ for all $m \in \{1 \dots 4\}$ and ranked the submitted systems based on this averaged value (\textit{RANK}). To better handle word duplicates, the computation is based on the position of words in a sentence rather than the actual words. 
Note that there were many cases where two or more tokens have the exact same probability. In this case, if the model predicts either one of the labels, we considered it as a correct answer.
Table \ref{tab:exp} shows some examples form the dataset, illustrating how the metric is computed. 

\section{Baseline Model}
We provided a baseline model for this task. This model (DL-BiLSTM-ELMo) is a sequence-labeling model that essentially utilizes ELMo contextualized embeddings \cite{peters2018deep} as well as two BiLSTM layers to label emphasis. During the training phase, the Kullback-Leibler Divergence (KL-DIV) ~\cite{kullback1951information} is used as the loss function. 
More analysis and the complete description of this model is provided in \cite{shirani2019learning}. 

\section{Systems and Results}
This task attracted 197 participants and a total of 31 teams made submissions to this task. The teams that submitted papers for the SemEval-2020 proceedings are listed in Table \ref{tab:leaderboard}. 
In total, 25 teams performed higher than the baseline and six teams performed lower.
13 of the 31 teams also submitted their system description papers.

\begin{figure}[h]
\centering
\includegraphics[width=0.50\linewidth]{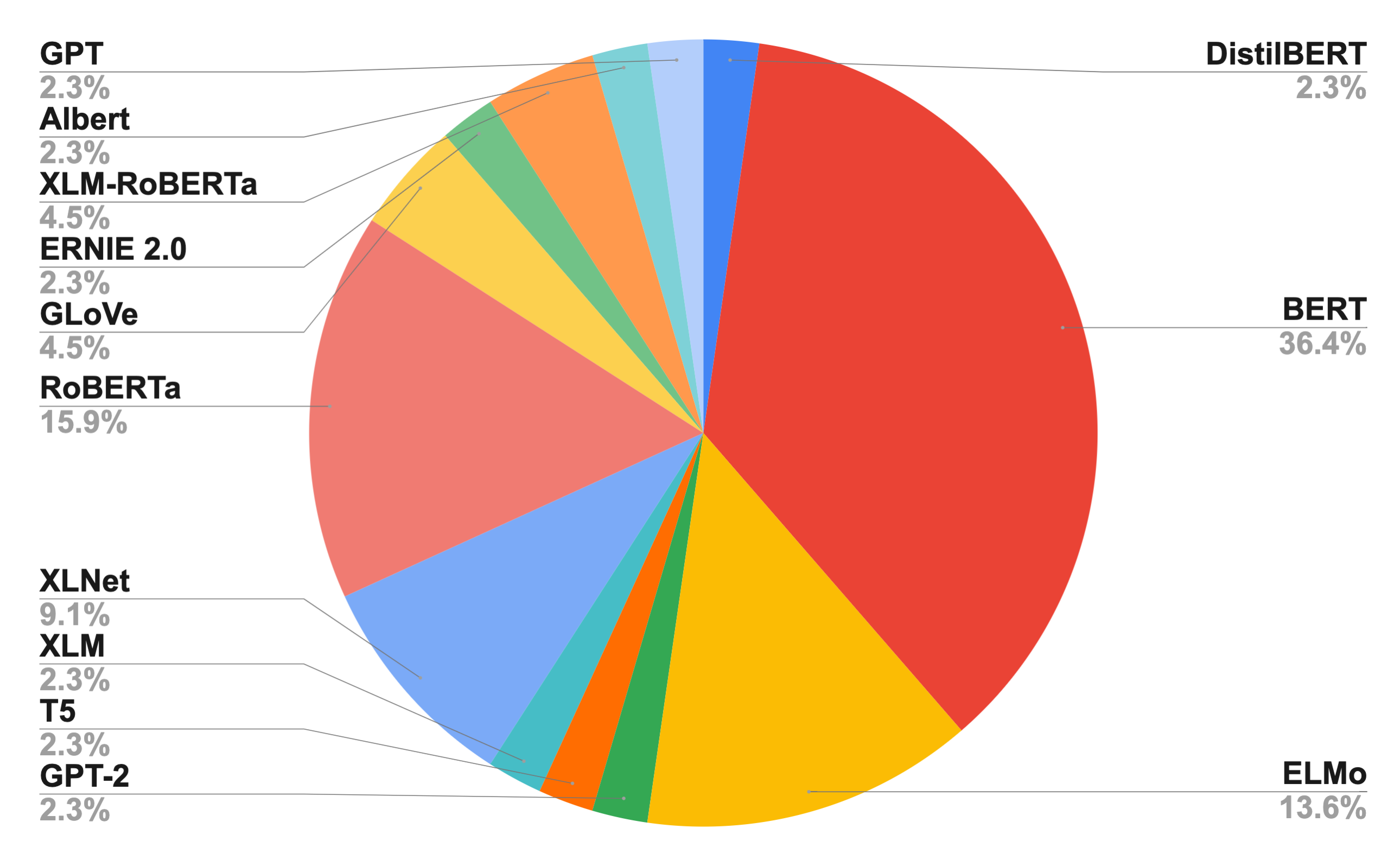} 
\caption{Pie chart showing the pre-trained models used in this task.}
\label{fig:pie}
\end{figure}

The base models used in the task submissions ranged from ELMo \cite{peters2018deep}, BERT \cite{devlin2018bert} and RoBERTa \cite{liu2019roberta}, to state-of-the-art pre-trained models such as ERNIE 2.0 \cite{sun2019ernie}, XLM-RoBERTa \cite{conneau2019unsupervised}, T5 \cite{raffel2019exploring} and ALBERT \cite{lan2019albert}. 
Figure \ref{fig:pie} shows different pre-trained models used in this task. Among them, BERT and RoBERTa were used most often. 
Ensemble transformer-based models were one of the most popular approaches (26\% of submissions).
All submissions applied deep neural network techniques to model emphasis. Moreover, some teams did explore hand-crafted features, such as part-of-speech tags, named entities, valence, arousal, dominance scores to enhance the performance of their models.

\subsection{Top Systems}
The results for each of four scores, as well as the \textit{RANK} score, are shown in Table \ref{tab:leaderboard}.
The top-3 teams based on the \textit{RANK} score are \textit{ERNIE} \cite{Huang2020ERNIE}, \textit{Hitachi} \cite{Morio2020Hitachi}, \textit{IITK} \cite{Singhal2020IITK}. 
The top-performing team, \textit{ERNIE}, achieved the highest $\mbox{Match}_m$ score of 0.823, 0.009 points higher than the second team and 0.013 points higher than the third team. 
\textit{ERNIE}, achieved the highest score not only in \textit{RANK} score but across all four scores. The next system on our leader board is \textit{Hitachi}, with a score of 0.814. And finally, \textit{IITK}, by achieving 0.810 \textit{RANK}, stands in third place.

\begin{table*}[h]
\caption{List of teams that participated in SemEval-2020 Task 10 with their ranks and scores.}
\label{tab:leaderboard}
\begin{tabular}{cccccccc}
\hline
\hline
\textbf{\#} & \textbf{Rank} & \textbf{Team Name} & \textbf{\textit{RANK} Score} & \textbf{Score 1} & \textbf{Score 2} & \textbf{Score 3} & \textbf{Score 4} \\ \hline
\textbf{1}  & 1             & ERNIE              & \textbf{0.823}               & \textbf{0.724} (1)        & \textbf{0.819} (1)        & \textbf{0.862} (1)        & \textbf{0.887} (1)        \\
\textbf{2}  & 2             & Hitachi            & 0.814               & 0.715 (2)        & 0.811 (3)        & 0.851 (6)        & 0.880 (3)        \\
\textbf{3}  & 3             & IITK   & 0.81                & 0.694 (5)        & 0.812 (2)        & 0.854 (3)        & 0.879 (4)        \\
\textbf{4}  & 4             & Randomseed19       & 0.805               & 0.677 (8)        & 0.803 (4)        & 0.858 (2)        & 0.881 (2)        \\
\textbf{5}  & 5             & Sherry             & 0.799               & 0.677 (8)        & 0.799 (5)        & 0.850 (7)        & 0.870 (7)        \\
\textbf{6}  & 5             & Sattiy             & 0.799               & 0.676 (9)        & 0.797 (6)        & 0.850 (7)        & 0.871 (6)        \\
\textbf{7}  & 6             & Unixlong             & 0.798               & 0.674 (10)        & 0.797 (6)        & 0.852 (5)        & 0.868 (8)        \\
\textbf{8}  & 6             & Giftai             & 0.798               & 0.674 (10)        & 0.793 (7)        & 0.853 (4)        & 0.870 (7)        \\
\textbf{9}  & 7             & FPAI             & 0.796               & 0.690 (7)        & 0.780 (10)        & 0.840 (9)        & 0.873 (5)        \\
\textbf{10}  & 7             & PAER             & 0.796               & 0.699 (3)        & 0.782 (8)        & 0.833 (12)        & 0.870 (7)        \\
\textbf{11}  & 8             & YYYY             & 0.795               & 0.694 (5)        & 0.780 (10)        & 0.835 (10)        & 0.871 (6)        \\
\textbf{12}  & 9             & L2020             & 0.794               & 0.696 (4)        & 0.781 (9)        & 0.831 (13)        & 0.868 (8)        \\
\textbf{13}  & 10             &BugHunter             & 0.785               & 0.654 (13)        & 0.775 (12)        & 0.845 (8)        & 0.867 (9)        \\
\textbf{14}  & 11             & Amobee             & 0.783               & 0.692 (6)        & 0.766 (16)        & 0.822 (17)        & 0.853 (17)        \\
\textbf{15}  & 11             & MIDAS             & 0.783               & 0.650 (16)        & 0.780 (10)        & 0.834 (11)        & 0.868 (8)        \\
\textbf{16}  & 12             & procyon             & 0.781               & 0.661 (11)        & 0.777 (11)        & 0.830 (14)        & 0.859 (12)        \\
\textbf{17}  & 12             & new\_bill\_\_             & 0.781               & 0.655 (12)        & 0.774 (13)        & 0.834 (11)        & 0.862 (11)    \\
\textbf{18}  & 12             & Jupyter             & 0.781               & 0.653 (14)        & 0.772 (14)        & 0.835 (10)        & 0.863 (10)    \\
\textbf{19}  & 13             & CrazyRock             & 0.774               & 0.651 (15)        & 0.766 (16)        & 0.824 (15)        & 0.857 (14)    \\
\textbf{20}  & 14             & CLP             & 0.772               & 0.642 (17)        & 0.763 (17)        & 0.823 (16)        & 0.858 (13)    \\
\textbf{21}  & 15             & TextLearner             & 0.767               & 0.627 (18)        & 0.769 (15)        & 0.823 (16)        & 0.850 (19)    \\
\textbf{22}  & 16             & Bright             & 0.758               & 0.627 (18)        & 0.749 (19)        & 0.809 (20)        & 0.848 (21)    \\
\textbf{23}  & 17             & LAST             & 0.756               & 0.610 (20)        & 0.749 (19)        & 0.812 (19)        & 0.853 (17)    \\
\textbf{24}  & 18             & AP            & 0.754               & 0.612 (19)        & 0.741 (20)        & 0.806 (22)        & 0.855 (16)    \\
\textbf{25}  & 18             & T\"extmarkers             & 0.754               & 0.596 (22)        & 0.752 (18)        & 0.815 (18)        & 0.851 (18)    \\
\textbf{\underline{26}}  & \underline{19}             & \underline{Baseline}    & \underline{0.75}   & \underline{0.608} (21)      & \underline{0.737} (21)  &\underline{0.807} (21)        & \underline{0.849} (20)    \\
\textbf{27}  & 20             & EL-BERT             & 0.745               & 0.591 (23)        & 0.726 (22)        & 0.807 (21)        & 0.856 (15)    \\
\textbf{28}  & 21             & David3             & 0.705               & 0.514 (24)        & 0.693 (23)        & 0.784 (23)        & 0.830 (22)    \\
\textbf{29}  & 22             & SRIB\_OnDeviceAI             & 0.659               & 0.479 (26)        & 0.640 (24)        & 0.732 (24)        & 0.785 (23)    \\
\textbf{30}  & 22             & UIC-NLP             & 0.659               & 0.490 (25)        & 0.638 (25)        & 0.731 (25)        & 0.775 (24)    \\
\textbf{31}  & 23             & IDS             & 0.629               & 0.425 (27)        & 0.618 (26)        & 0.716 (26)        & 0.756 (25)    \\
\textbf{32}  & 24             & YNU-HPCC             & 0.401               & 0.231 (28)        & 0.373 (27)        & 0.467 (27)        & 0.534 (26)    \\
\hline
\hline
\end{tabular}
\end{table*}

\begin{table}[]
\centering
\caption{Main features in participating systems. An `N/A' in the \textbf{Ref.} column means that we did not receive a system description paper for that entry.}
\label{tab:methods}
\resizebox{\textwidth}{!}{%
\begin{tabular}{l|l|l|l}
\hline
\textbf{Rank} & \textbf{Team Name} & \textbf{Ref.} & \textbf{Best performing Methods} \\ \hline\hline
1 & ERNIE & \cite{Huang2020ERNIE} & ERNIE 2.0 + data augmentation + hand-crafted features \\ \hline
2 & Hitachi & \cite{Morio2020Hitachi} & \begin{tabular}[c]{@{}l@{}}Distribution fusion of BERT, GPT-2, RoBERTa,\\ XLM-RoBERTa, XLNet, XLM, T5\\ + POS tags and token embeddings\end{tabular} \\ \hline
3 & IITK & \cite{Singhal2020IITK} & Ensemble of BERT, RoBERTa and XLNet \\ \hline
4 & Randomseed19 & \cite{Shatilov2020randomseed} & Ensemble of BERT, RoBERTa and XLNet \\ \hline
5 & Sherry & N/A & \begin{tabular}[c]{@{}l@{}}Ensemble of Bert, XLM-RoBERTa and RoBERTa\\ + POS tags for adjusting final outputs\end{tabular} \\ \hline
5 & Sattiy & N/A & \begin{tabular}[c]{@{}l@{}}Aggregation of BERT, XLNet and GPT\\(formulated the task as a 10-class classification task) \end{tabular} \\ \hline
7 & FPAI & \cite{Guo2020FPAI} & \begin{tabular}[c]{@{}l@{}}RoBERTa-base (converted the task to a query-based\\ machine reading comprehension task)\end{tabular} \\ \hline
10 & BugHunter & N/A & \begin{tabular}[c]{@{}l@{}}LSTM model using BERT, POS tags and\\ tf-idf features (multi-task framework with\\ classification and distribution training objectives)\end{tabular} \\ \hline
11 & Amobee & N/A & Combination of BERT and RoBERTa \\ \hline
11 & MIDAS & \cite{Gupta2020MIDAS} & Ensemble of RoBERTa, BERT and XLNet \\ \hline
12 & Procyon & N/A & \begin{tabular}[c]{@{}l@{}}ELMo (multi-modal model based on\\ word representations and POS tags)\end{tabular} \\ \hline
12 & Jupyter & N/A & \begin{tabular}[c]{@{}l@{}}BERT + sentiment values such as pleasantness,\\ attention, sensitivity, aptitude and polarity\end{tabular} \\ \hline
13 & CrazyRock & N/A & BERT + 4 parallel Bi-GRU + 2 layers of self attention \\ \hline
14 & CLP & N/A & \begin{tabular}[c]{@{}l@{}}BERT + BiLSTM + hand-crafted features\\ (listwise ranking model)\end{tabular} \\ \hline
15 & TextLearner & \cite{Yang2020TextLearner} & RoBERTa (two-staged ranking model) \\ \hline
16 & Bright & N/A & BERT and ELMo (Graph neural networks) \\ \hline
17 & LAST & \cite{Bestgen2020LAST} & BERT and ELMo with LightGBM \\ \hline
18 & AP & N/A & ELMo with BiLSTM \\ \hline
18 & T\"extmarkers & \cite{Glocker2020Textmarkers} & BERT, GRU, attention + crowd layer \\ \hline
20 & EL-BERT & \cite{Kanani2020ELBERT} & ELMo + POS + SentBERT + WordBERT \\ \hline
22 & UIC-NLP & \cite{Hossu2020UICNLP} & \begin{tabular}[c]{@{}l@{}}GloVe, LSTM, POS and valence, arousal,\\ dominance (VAD) scores\end{tabular} \\ \hline
23 & IDS & \cite{Shin2020IDS} & \begin{tabular}[c]{@{}l@{}}BERT, DistilBERT, GPT-2, RoBERTa, XLNet, and XLM\\ (using self-attention distributions of PLMs\\ in a zero-shot setting)\end{tabular} \\ \hline
24 & YNU-HPCC & \cite{Shin2020YNUHPCC} & \begin{tabular}[c]{@{}l@{}}ELMo and BERT (multi-granularity ordinal\\ classification model)\end{tabular} \\ \hline
\end{tabular}%
}
\end{table}

\subsection{Top Performing Systems and Novel Architectures }
In this section, we provide a brief description of the best performing and novel approaches. Table \ref{tab:methods} shows a high level summary of these systems.

\textit{ERNIE} achieved the highest score by fine-tuning ERNIE 2.0 as the base model. They also reported high performance by using other pre-trained models like XLM-RoBERTa, RoBERTa and ALBERT. They further boosted the model by utilizing data augmentation and hand-crafted features like word capitalization and the occurrence of hashtags in instances.  

\textit{Hitachi} tackled the task by combining rich contextualized embeddings and fine-tuning seven Pre-trained Language Models (PLMs) on the task such as BERT, GPT-2 \cite{radford2019language}, RoBERTa, XLM-RoBERTa, XLNet \cite{yang2019xlnet}, XLM \cite{lample2019cross}, and T5. In addition, they added POS tags and token embeddings from a character-level LSTM layer. 
They introduced a distribution fusion system to fuse the output distributions of the fine-tuned models and find the optimal hyperparameter set. They showed the performance gain of the fusion model over average ensemble as well as individual PLMs. 
Among all PLMs, BERT and XLNet models were more successful in predicting emphasis individually. 

\textit{IITK}, the team ranking in third place, proposed an ensemble model where the base models were BERT, RoBERTa, and XLNet. In order to aggregate the outputs, they computed the average of the scores predicted by these models.
The authors also provided different baselines from the character-level BiLSTM model with attention to transformer-based models like XLM-RoBERTa, ALBERT and GPT-2. When comparing all individual models, XLNet-Large performed the best. 

A wide range of novel methods were used to model emphasis. For example, 
\textit{FPAI} \cite{Guo2020FPAI} converted the task of emphasis selection to a simplified query-based machine reading comprehension (MRC) task, where the goal was to answer the fixed query, ``Find candidates for emphasis”. 

To tackle the low inter-annotator agreement in the dataset, \textit{T\"extmarkers} \cite{Glocker2020Textmarkers} attempted to model multiple annotators jointly by adapting a crowd layer architecture \cite{rodrigues2018deep}, introducing initialization with agreement dependent noise. The crowd layer is intended to help the model to outperform a baseline trained with token level majority voting.

\textit{IDS} \cite{Shin2020IDS} performed an interesting analysis of pre-trained models to investigate whether PLMs contain enough knowledge to select proper words for emphasis.
They compared different zero-shot models in which self-attention distributions of PLMs were used to emphasize words.
More precisely, the authors investigated individual attention heads of different models like BERT, DistilBERT, GPT-2, RoBERTa, XLNet, and XLM to probe their ability to identify emphasis without any fine-tuning. Their interesting findings indicate that DistilBERT is more successful in predicting emphasis while XLNet and GPT-2 perform poorly when there is no training for this task. 

The top non-transformer-based model, \textit{Procyon} (ranked 12th), successfully proposed an ELMo-based multi-modal model with two sub-networks to learn emphasis scores based on word representations and POS tags separately. 

\subsection{Discussion}
To have a better understanding of the challenges of this task, we perform an error analysis to examine where the models succeed and in what situations they face difficulties in selecting emphasis words.
More specifically, we compute the average $\mbox{Match}_m$ score over all 31 submissions for every example in the test set and examine the challenging cases for all models. 
Table \ref{tab:exp} shows some interesting examples from the test set with three $\mbox{Match}_m$ scores ($m1$--$m3$) from all submissions, where $m1$ stands for the average score for system predictions obtained by selecting the top word, and $m3$ stands for results from selecting the top 3. 
In many cases, selecting emphasis words was unchallenging for most of the systems (e.g., S1 in Table \ref{tab:exp} with ``Imagination" as the top word and ``rules" and ``world" with same emphasis probability.).
In some examples, there is no single token standing out in the sentence, so it was not easy to select one single word with certainty. S2 is a good example with low $m1$ and high $m2$ and $m3$, indicating disagreement between models and annotators for choosing the first word with the highest probability.
We also observed many cases where one word clearly stands out of the sentence but it is not clear which words should be selected next. S3 is an example of this where most systems were able to select the top word ``talked" correctly, but faced difficulties in predicting other words for that sentence.

There are some cases where prediction is easy for humans but still poses challenges for models. For example, most annotators agreed on selecting ``basketball" with the highest probability in S4; however, many models failed to select this word in the top position, probably due to the unusual structure of the sentence. In this example, ``East", ``Sleep" and ``Watch" have equal probabilities in the annotation.

\begin{table*}[h]
\centering
\caption{Examples from the test set with averaged $\mbox{Match}_m$ scores across all submitted systems. Words with high emphasis probability labels are shown in bold. 
}
\label{tab:exp}
\begin{tabular}{l|l|l}
\hline 
\multicolumn{1}{c|}{\textbf{Num}}  & \textbf{Sentence}                                                                                                           & \multicolumn{1}{c}{\textbf{$\mbox{Match}_m$}} \\ \hline\hline
\multicolumn{1}{c|}{\textbf{S1-1}} & \textbf{Imagination} rules the world.                                                                                               & \multicolumn{1}{c}{$m1$ = 0.9354}     \\ \hline
\textbf{S1-2}                      & \textbf{Imagination} \textbf{rules} the \textbf{world}.                                                                                               & $m2$ = 1.0                            \\ \hline
\textbf{S1-3}                      & \textbf{Imagination} \textbf{rules} the \textbf{world}.                                                                                               & $m3$ = 1.0                            \\ \hline
\multicolumn{1}{c|}{\textbf{S2-1}} & \begin{tabular}[c]{@{}l@{}}All successes begin with self-discipline. It starts with \textbf{you}.\end{tabular}                    & \multicolumn{1}{c}{$m1$ = 0.0322}     \\ \hline
\textbf{S2-2}                      & \begin{tabular}[c]{@{}l@{}}All successes begin with \textbf{self-discipline}. It starts with \textbf{you}.\end{tabular}                    & $m2$ = 0.8225                         \\ \hline
\textbf{S2-3}                      & \begin{tabular}[c]{@{}l@{}}All \textbf{successes} begin with \textbf{self-discipline}. It \textbf{starts} with \textbf{you}.\end{tabular}                    & $m3$ = 0.9677                         \\ \hline
\multicolumn{1}{c|}{\textbf{S3-1}} & \begin{tabular}[c]{@{}l@{}}I learned most, not from those who taught me\\ but from those who \textbf{talked} with me.\end{tabular} & \multicolumn{1}{c}{$m1$ = 0.8387}     \\ \hline
\textbf{S3-2}                      & \begin{tabular}[c]{@{}l@{}}I learned most, not from those who taught me\\ but from those who \textbf{talked} \textbf{with} me.\end{tabular} & $m2$ = 0.5645                         \\ \hline
\textbf{S3-3}                      & \begin{tabular}[c]{@{}l@{}}I learned most, not from those who taught me\\ but from those who \textbf{talked} \textbf{with} \textbf{me}.\end{tabular} & $m3$ = 0.4516                         \\ \hline
\multicolumn{1}{c|}{\textbf{S4-1}} & Eat . Sleep . Watch \textbf{Basketball} . Repeat .                                                                                   & \multicolumn{1}{c}{$m1$ = 0.1290}     \\ \hline
\textbf{S4-2}                      & \textbf{Eat} . \textbf{Sleep} . \textbf{Watch} \textbf{Basketball} . Repeat .                                                                                   & $m2$ = 0.6935                         \\ \hline
\textbf{S4-3}                      & \textbf{Eat} . \textbf{Sleep} . \textbf{Watch} \textbf{Basketball} . Repeat .                                                                                   & $m3$ = 0.7419                         \\ \hline
\end{tabular}
\end{table*}
\section{Conclusion}
This paper summarizes the insights gained from organizing Task 10 at SemEval-2020. Given a short piece of text, the task consists of selecting candidate words to emphasize. We received a good number of system submissions, with 13 teams submitting a system description paper. While there were many differences between individual systems, we observed a strong trend favoring the use of transformer based models as key ingredient in the proposed architectures. Many description papers present valuable analyses of the data and task. We encourage readers interested in this task to take a careful look at these papers for additional inspiration on how to improve results further.



\bibliographystyle{coling}
\bibliography{semeval2020}

\end{document}